\documentclass[twoside,11pt]{article}

\usepackage[abbrvbib]{jmlr2e}
\usepackage{sub caption}
\usepackage{tikz}
\def\checkmark{\tikz\fill[scale=0.4](0,.35) -- (.25,0) -- (1,.7) -- (.25,.15) -- cycle;} 

\ShortHeadings{ShortFuse: Learning Time Series Representations with Structured Information}{Fiterau, Bhooshan, Fries, Bournhonesque, Hicks, Halilaj, R\'{e} and Delp}

\firstpageno{1}

\begin{document}

\title{ShortFuse: Biomedical Time Series Representations in the Presence of Structured Information}

\author{\name Madalina Fiterau \email mfiterau@cs.stanford.edu \\
            \addr Computer Science Department, Stanford University
            \AND
            \name Suvrat Bhooshan \email suvrat@stanford.edu \\
            \addr Computer Science Department, Stanford University
            \AND  
            \name Jason Fries \email jason-fries@stanford.edu \\
            \addr Computer Science Department, Stanford University
            \AND 
            \name Charles Bournhonesque \email cbournho@stanford.edu \\
            \addr Institute for Computational and Mathematical Engineering, Stanford University
            \AND
            \name Jennifer Hicks \email jenhicks@stanford.edu \\
            \addr Bioengineering Department, Stanford University
            \AND
            \name Eni Halilaj \email ehalilaj@stanford.edu \\
            \addr Bioengineering Department, Stanford University
            \AND
            \name Christopher R\'{e} \email chrismre@cs.stanford.edu \\
            \addr Computer Science Department, Stanford University
            \AND
            \name Scott Delp \email delp@stanford.edu \\
            \addr Bioengineering Department, Stanford University
}

\maketitle

\begin{abstract}
In healthcare applications, temporal variables that encode movement, health status and longitudinal patient evolution are often accompanied by rich structured information such as demographics, diagnostics and medical exam data. However, current methods do not jointly optimize over structured covariates and time series in the feature extraction process. We present ShortFuse, a method that boosts the accuracy of deep learning models for time series by explicitly modeling temporal interactions and dependencies with structured covariates. ShortFuse introduces hybrid convolutional and LSTM cells that incorporate the covariates via weights that are shared across the temporal domain. ShortFuse outperforms competing models by 3\% on two biomedical applications, forecasting osteoarthritis-related cartilage degeneration and predicting surgical outcomes for cerebral palsy patients, matching or exceeding the accuracy of models that use features engineered by domain experts. 
\end{abstract}

\section{Introduction}
\label{sec:intro}

In biomedical applications, time series data frequently co-occur with structured information. These time series data vary widely in form and temporal resolution, from high-frequency vital signs to longitudinal health indicators in an electronic medical record to activity monitoring data recorded by accelerometers. Structured covariates, such as patient demographics and measures from clinical examinations, are common and complementary to these time series data. While large amounts of these types of data are available, they are in many cases challenging to integrate and analyze.

For instance, consider data from patients with cerebral palsy (CP), a condition that affects approximately 3 out of every 1000 children in the U.S. \citep{bhasin2006prevalence}. Cerebral palsy makes walking inefficient and sometimes painful. Musculoskeletal surgeries can improve walking, but outcomes are highly variable. Extensive data is available to aid treatment planning, including gait analysis data that characterizes the motion of each joint (e.g., hip, knee, and ankle) during gait, along with a host of structured data such as strength and flexibility measures and birth history (e.g., number of weeks born premature). At many clinical centers, there are roughly as many structured covariates as time series features, from high resolution gait data to clinical visit records kept over several years. All these interconnected factors make treatment planning difficult.

Current methods for analyzing these types of data rely on extensive feature engineering, often modeling the time-series and structured information independently. Standard transformations such as Principal Component Analysis (PCA) can be insufficient for capturing all information in time series, requiring additional feature engineering by domain experts. Traditionally, when methods such as PCA, Multiple Kernel Learning (MKL), Dynamic Time Warping (DTW), neural networks or other transformations are used to extract features from time series, the structured covariates in the datasets have no impact on the learned temporal features. In most biomedical applications, there are interactions and correlations between time series and covariates that we would like to leverage. In the case of cerebral palsy, younger children or those with a more severe neural injury might have different gait features that help predict an appropriate surgical plan.

To address this issue, we introduce ShortFuse, a method that boosts the accuracy of deep learning models for time series by leveraging the structured covariates in the dataset. The key to learning relevant representations is to take into account the specifics of the covariates. For instance, cerebral palsy subjects are seen and treated from toddlerhood to adulthood and the temporal patterns in the joint motion waveforms will depend on the subject's stage of development -- while walking on the toes is normal in toddlers, it is an abnormality in older children and adults. By definition, the structured information for a sample is constant along the temporal domain, which is why its corresponding parameters should not be allowed considerable variations, as that would translate to an additional intercept term and result in overfitting. As an illustration of this, consider that two successive gait cycles for the same subject could result in vastly different representations by tweaking the time-varying weights of the covariates, which is why this temporal variation of the weights should be discouraged through parameter sharing. Finally, as clinicians try to keep as complete records as possible, there might be numerous fields denoting clinical tests and patient history which, unlike the time series data, might not be relevant to the predictive task at hand, so there must be some mechanism to discount them and not explode the parameter space.

ShortFuse preserves the sequential structure of time series, explicitly modeling interactions and dependencies with structured covariates, allowing the latter to guide feature learning and improve predictive performance. Our approach introduces specialized structures, which we call `hybrid layers' for fusing structured  covariates with time series data. The hybrid layers use structured information as distinct inputs, which are used to parametrize, guide, and enrich the feature representations. The first type of layer uses convolutions parametrized by the covariates, where the weights of the structured covariates are shared across the convolutions. Secondly, we introduce an LSTM hybrid, which shares the covariates and their weights across the cells and uses them in the computation of the input gate, forget gate, state change and output layer. The LSTM hybrid is thus able, for instance, to adjust the length of the forget window according to the structured information.

We demonstrate, through two representative biomedical applications, the versatility of ShortFuse, which makes no assumptions regarding the structure, dimensionality, or sampling frequency of the time series. We also show that the method is flexible, in that it can be applied to RNN or CNN model architectures. Through the two biomedical examples, we show that adding structured covariates boosts the accuracy of a time-series-only deep learning model. In addition, ShortFuse matches or improves on results obtained through feature engineering performed by domain experts, achieving state-of-the-art accuracy with no manual feature engineering. While the focus here is on biomedical applications, ShortFuse is also applicable to vehicle monitoring, financial forecasting, activity recognition based on sensor arrays, and prediction of cyberattacks from net traffic.

\section{Related Work}
\label{sec:related}
Several different approaches have been used to featurize time series for integration with structured information. A simple approach is to construct  histograms of the values in the sequence and operate solely on count data. This is a common approach to extract features about physical activity intensity from accelerometer data, e.g. \citep{dunlop2011objective,song2010assessing}. Principal Component Analysis (PCA) can summarize signals by extracting the linear combinations that account for most of the variance in the data. For example, previous investigators have used PCA to extract features from joint motion waveforms measured during walking and running and then appended the principal components to other structured information, e.g. \citep{astephen2008gait,federolf2013application}. Segmentation of periodic signals into periodic intervals is also widely used for the processing of vital signs such as ECG \citep{keogh2001online,keogh2004segmenting} to extract features such as peak-to-peak variability. Methods that account for time series similarity, such as Dynamic Time Warping (DTW) were previously applied to sensor data from an Inertial Measurement Unit (IMU) for gait recognition \citep{kale2003gait} in combination with age and gender information \citep{trung2012performance} and the study of gait in subjects with Parkinson's disease \citep{wang2016monitoring}. Other sophisticated methods such as Multiple Kernel Learning (MKL) \citep{aiolli2014easy} have been applied, for example, to classify, based on electroencephalography (EEG) signals, which parts of the brain are important for subjects performing gait movements (stop, walk, turn) \citep{zhang2017multiple}. Hand engineered features, such as summary statistics, ranges of values, and spectral data extracted from the signal are also frequently employed (e.g., for accelerometer-based activity data \citep{lee2015sedentary} and joint motion waveforms \citep{truong2011evaluation,fukuchi2011support}. All these existing methods for combining covariates with time series are highly specialized for their intended application. Optimizing feature representations over all data, temporal and structured, could improve predictive performance by accounting for the interdependence between temporal and structured data.

Deep learning obviates this need for feature engineering and provides a general method to integrate time series and structured covariates, but approaches of joint optimization over these data  are largely unexplored. In the past, CNNs and LSTMs have proven apt at encoding temporal information. RNN and LSTMs were shown to perform well for vital signs \citep{graves2013speech}. Deep CNNs perform well for network traffic monitoring \citep{wang2016earliness}, financial time series \citep{borovykh2017conditional}, audio \citep{zheng2014time} and clinical diagnostics \citep{RazavianS15}. Multiscale or Multiresolution CNNs were recently shown to perform well on time series benchmark tasks \citep{cui2016multi}. Encoders can improve the performance of architectures, for example in the case of anomaly detection in vehicle sensors \citep{malhotra2016lstm}. 

Given the wide range of available deep learning architectures, one could trivially introduce covariates at the bottom levels of existing models by replicating them along the temporal dimension, thus obtaining a constant sequence for each covariate, which can be added to existing time series. This poses multiple problems. LSTM layers learn from variations along the temporal domain, which do not appear in covariates. With convolutions, there is no parameter sharing, which means the replicated covariates are treated as separate inputs, which can easily lead to overfitting due to the introduction of parameters for each of the covariates at each time point, which are not needed as the covariates themselves do not change in time. Also, there are no shortcut connections that may link covariates to later stages in the network, which restricts the information flow. This misses the opportunity for something akin to skip connections \citep{sermanet2013pedestrian}, which can enrich representations by connecting arbitrary levels in the network.  ShortFuse overcomes all these issues by jointly learning representations over heterogeneous time series and structured data though the hybrid layers described in detail in the next section.

\section{ShortFuse}
\label{sec:shortfuse}

In the sections below, we discuss the fusion of information from time series and structured data using deep neural networks, and introduce the technical contributions of ShortFuse.

\subsection{Information Fusion}
\label{sec:fusion}

First, we discuss the cases when fusing time series data and covariates leads to improved predictive performance. We assume that the input data has a set of $d$ structured covariates. $S$ is the design matrix, the structured information in the dataset. $X$ is a fixed-length multivariate time series. We refer to the temporal variables as sequences or signals. $Y$ is the univariate output. For a sample $i$, we use $s_i$, $x_i$ and $y_i$ to indicate the covariate vector, time series and label, respectively. $x_i$ is a matrix of $n$ by $t$, where $n$ is the recorded number of sequences, or time series signals, and $t$ is the number of points in time at which the records were captured. $y$ is an integer representing the class label. 

Given that the covariates and the sequences typically record different clinical data, it is expected that a predictive model of $Y$ using both $X$ and $S$ will perform better than using either $X$ or $S$. A simple test to check whether the covariates contain additional information is that $Y$ is not conditionally independent of $S$ given $X$. In this case, $I(Y;S|X)\not=0$. Recent work in nonparametric estimation of mutual information \citep{reddi2013scale} makes it possible to perform this test. Similarly, if $I(Y;X|S)\not=0$, the covariates are also insufficiently informative. In other words, fusion of time series and covariates should be used when both of the conditional mutual information values are above $0$.

The simplest approach to introducing covariates in the deep learning model is to replicate each covariate, and provide them as inputs appended to the time series. Alternatively, they could be introduced in one of the intermediate layers or only used in the final layer in the network. These choices impact the learned features, as the relevant temporal features often depend directly on the structured covariates.

For instance, consider the case of two subjects with osteoarthritis, but different body mass index (BMI) values -- one subject is obese and one is normal weight. The task is to determine whether osteoarthritis will progress in time given the subjects' physical activity data, time-series data from accelerometers, and other structured covariates. For the obese subject, the important physical activity features may be different. For example, given the obese subject's higher weight to height ratio, intense activity may cause detrimental loading of the joint, while in the normal subject the same types of activity may not contribute to disease progression. Instead, for the healthy subject, the mean or minimum activity intensity is possibly more predictive. Thus, structured information present in the dataset has a direct impact on which features should be learned by the model. In this case, not considering the covariates runs the risk of producing less informative features.

\begin{figure}[htbp]
  \centering 
  \includegraphics[width=0.8\textwidth]{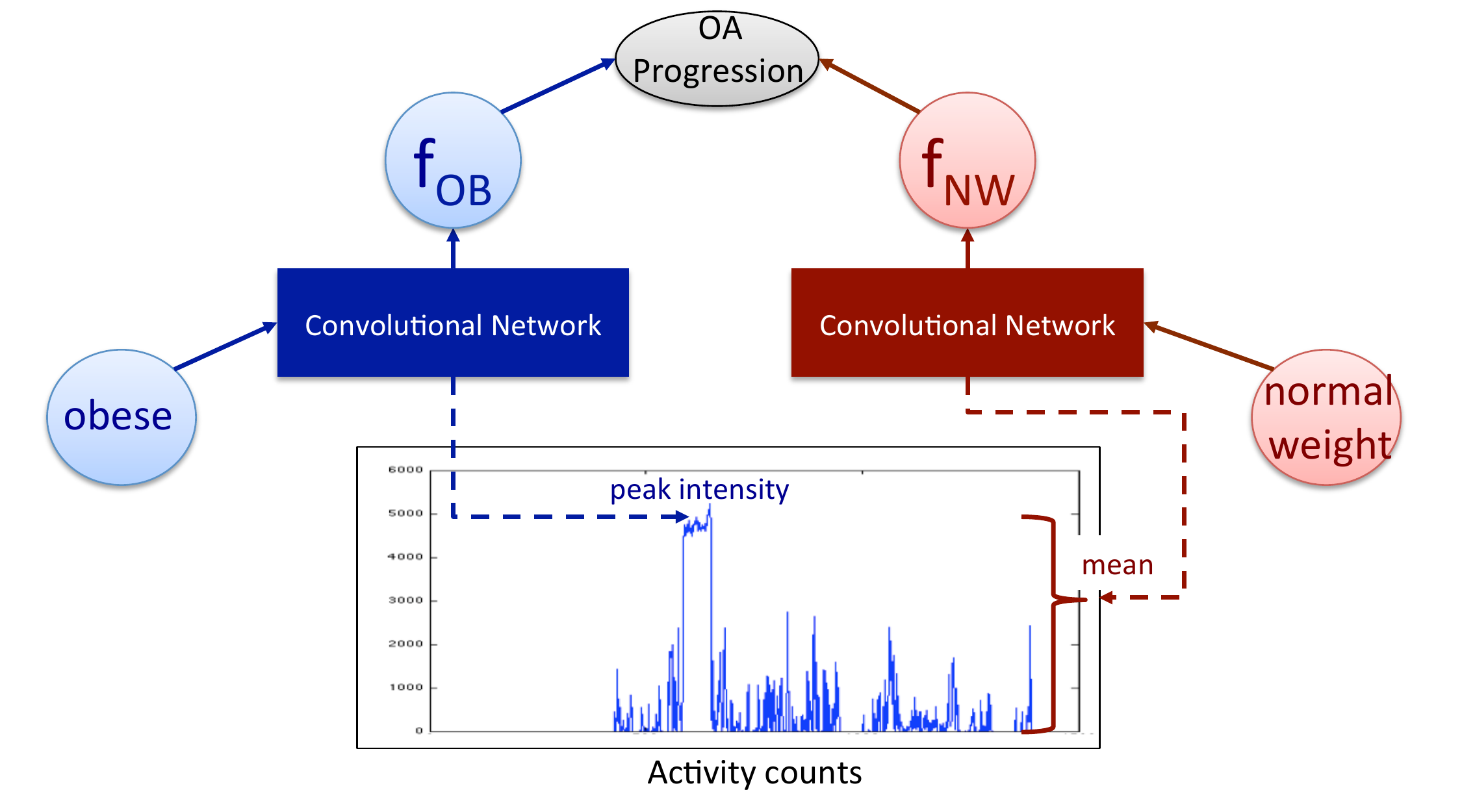} 
  \caption{Feature learning mechanism in the presence of covariates. $f_{OB}$ and $f_{NW}$ are internal representations learned by the network. The two parts of the convolutional network learn features relevant for obese subjects and subjects with normal weight, respectively.}
\label{fig:fusion}
\end{figure}

Figure~\ref{fig:fusion} illustrates a minimal structure that is capable, via back-propagation, of leveraging these dependencies to learn and use internal representations as appropriate for the predictive task. Assume there are two binary structured covariates,  `obese' and `normal weight', representing subjects' BMI status. If `obese' is active for a subject we'll use their data to update the first feature $f_{OB}$. Over time, the feature will become informative in determining whether the subjects osteoarthritis will progress for obese subjects. For instance, it might learn to encode maximum activity intensity. As the covariate `obese' is 0 for subjects with normal weight, feature $f_{OB}$ does not contribute to their predictions. The second feature $f_{NW}$ will be updated in the same way, using data from the subjects with a normal BMI. Overall, the internal representations are only influenced by the samples for which these representations are relevant. 

\subsection{Fused Architectures}
\label{sec:architectures}

ShortFuse works on the premise that the earlier the covariates are introduced into a network, the more they will be able to direct feature construction. Given a deep network designed for time series, ShortFuse constructs hybrid layers that use the covariates in such a manner that the representational capabilities of LSTMs and CNNs are preserved, meaning that the hypothesis space for the learned features is expanded. The key novelty of the hybrid layers is the treatment of structured covariates as global features that are combined with the local temporal patterns encoded by the network. The following section details the hybrid CNNs and LSTMs used to obtain our results, while a complete list, based on commonly used deep learning layers, is summarized in Appendix~A, Table~\ref{tab:hybrids}.

\subsubsection{Hybrid Convolutions}
\label{sec:hybridconv}

\begin{figure}[!htbp]
  \centering 
  \includegraphics[width=\textwidth]{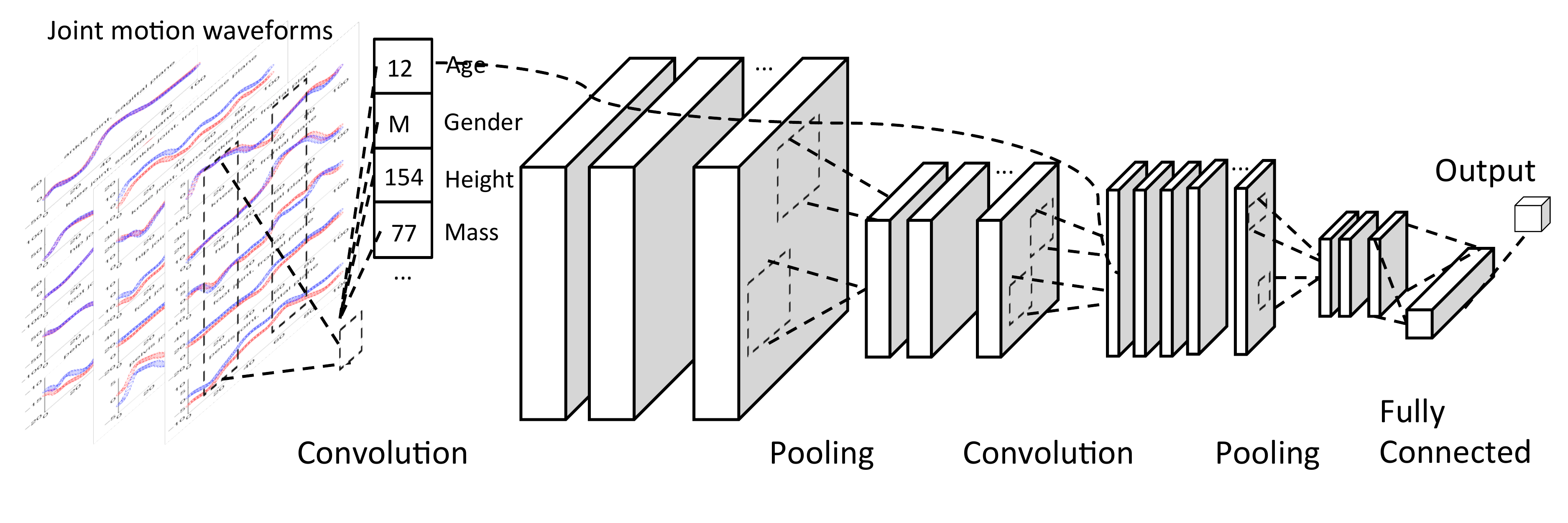} 
  \caption{Hybrid convolutional layers using structured covariates. The time series data is shown on the left. The convolutions, using dropout, are then applied to the sequences in a time window with the covariates (age, gender, height, and mass) as parameters. There can be several convolutional filters, the output of which are pooled, followed by another layer of convolutions which can, in turn, use the covariates. In this example, there is a second pooling layer followed by a fully connected layer and a softmax.}
\label{fig:conv}
\end{figure}

The ShortFuse hybrid convolutional layers, used predominantly though not exclusively in the initial layers of the network, provide the covariates as parameters to every convolution function along the temporal dimension, together with the time series in that specific window. Figure~\ref{fig:conv} shows a network with two convolutional hybrid layers. Dropout is used such that not all the covariates are provided to all the convolutions.

%For network architectures where the first layer is not a convolution, such as in the case of an embedding followed by convolutions, the covariates are introduced right before the first convolutional layer and merged with the time series transformations obtained so far. 

\subsubsection{Hybrid LSTMs}
\label{sec:hybridLSTM}

\newcommand\W[1]{W\textsubscript{#1}}

\begin{figure}[!htbp]
  \centering 
  \includegraphics[width=0.8\textwidth]{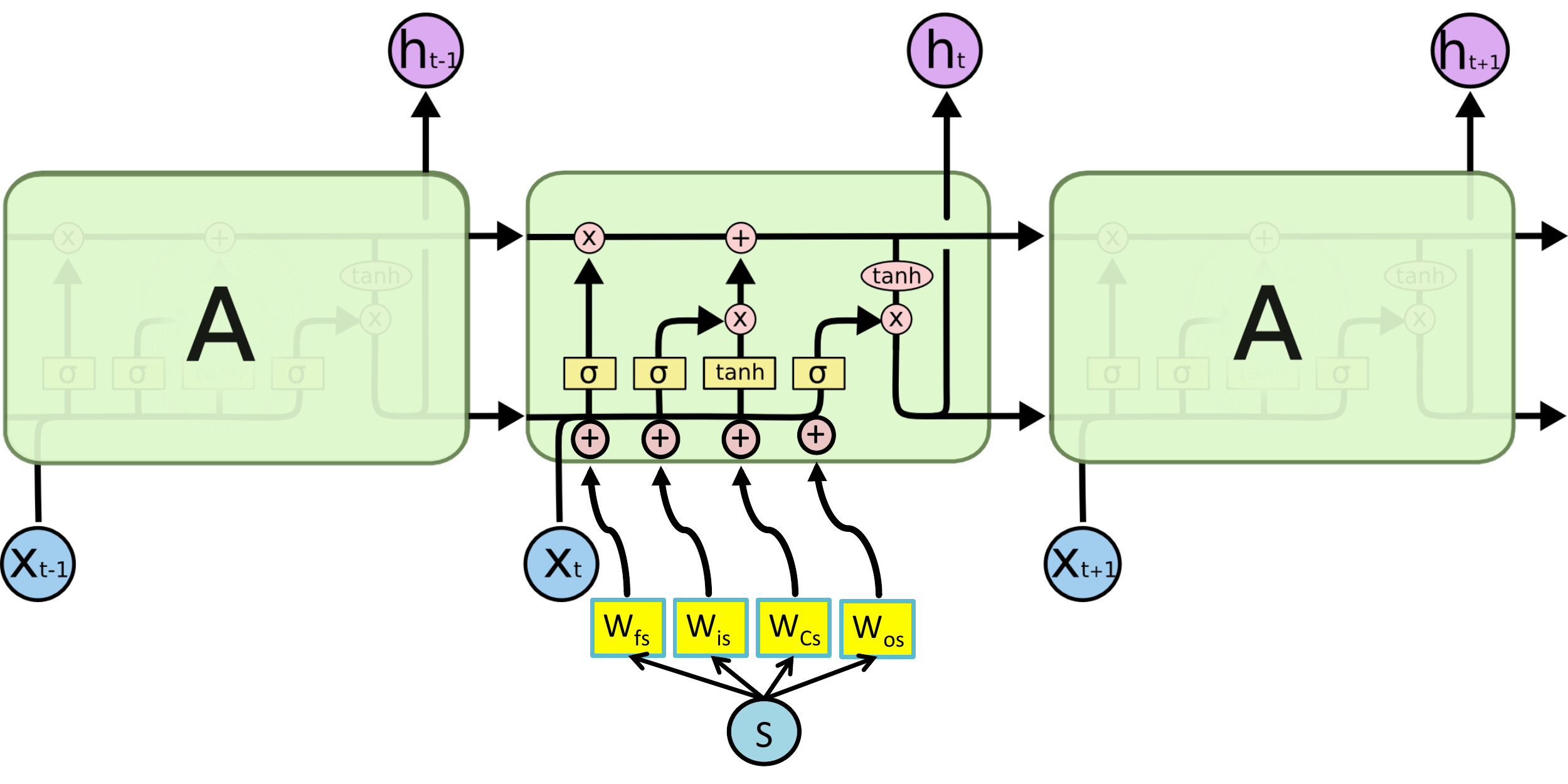} 
  \caption{Hybrid LSTM layer. The structured covariate weights are shown in yellow. A dot product between these parameters, shared across cells and the structured covariates is added to the original input to the LSTM nonlinearities. Within a cell, the symbols $+$ and $\times$ in the small circles represent binary operations, while $\tanh$ in the oval is the activation applied to the output. The functions in the yellow rectangles, $\sigma$, $\sigma$, $\tanh$ and $\sigma$ represent the nonlinearities of the LSTM for the forget gait, input gate, state change and output gate respectively. The outputs of the LSTM, $h_{t-1}$, $h_{t}$ and $h_{t+1}$, are the learned representations.}
\label{fig:LSTM}
\end{figure}

For LSTM-based architectures, the structured covariates are used internally by the LSTM as part of additive terms in the computation of the LSTM's nonlinearities, as shown in Figure~\ref{fig:LSTM}. We introduce weights $\W{fs}$(forget gate), $\W{is}$ (input gate), $\W{Cs}$(state change), $\W{os}$ (output gate). The added `s' in the subscript indices of the weights indicate that these weights correspond to the structured covariates $s$. The terms $\W{fs}\cdot s$, $\W{is}\cdot s$, $\W{Cs}\cdot s$ and $\W{os}\cdot s$ are  added to the arguments of each of the four nonlinearities in the LSTM. The time series values $x_{t-1}$, $x_t$ and $x_{t+1}$ are provided as input to the cells. The structured covariates $s$ for a given sample are shared across the LSTM cells, together with the covariate weights  $\W{fs}$, $\W{is}$, $\W{Cs}$ and $\W{os}$.

\subsubsection{Late Fuse}
\label{sec:latefuse}
A simple alternative to merging time series and covariates data uses a CNN on the structured covariates before the output layer (softmax, binary cross-entropy) of the network. The method is called LateFuse as the covariates are only considered at the end. LateFuse merges the outputs from the network on the time series data and from the covariate CNN.

\section{Experimental Design}
\label{sec:exp}

We developed an evaluation framework to compare ShortFuse (i.e., early use of covariates) to deep learning models that do not use covariates, along with LateFuse (features included in the top level), and methods which train classifiers on separate time series representations appended to structured covariates. We selected two representative biomedical applications: predicting good candidates for surgical treatment of cerebral palsy-related gait disorders and predicting cartilage degeneration in patients at risk for osteoarthritis.

\subsection{Candidate Models}
\label{sec:candidates}

For each application we tested several deep learning models that have been shown to perform well for time series, as discussed in Section~\ref{sec:related}. The contenders include an LSTM (Appendix~B, Figure~\ref{fig:4a}) \citep{graves2013speech}, a Deep CNN \citep{conneau2016very}, a Multiresolution CNN \citep{cui2016multi}, and a CNN network with an Encoder (Appendix~B, Figure~\ref{fig:4b}) \citep{malhotra2016lstm}. We compared these CNN and LSTM based models, following hyper-parameter tuning according to the application, against ShortFuse and LateFuse, using the top-performing deep learning model for each of the applications. 

We also ran Multiple Kernel Learning (MKL) and Dynamic Time Warping (DTW). MKL is a class of algorithms which uses linear combinations of a few predetermined kernels to predict the output \citep{aiolli2014easy}. Lambda is a regularization hyperparameter of the MKL algorithm which represents the minimizer of the 2-norm of the vector of distances. DTW is a similarity measure which computes the distance between two time-series. We predict our output using nearest neighbours wherein DTW is used as the distance measure between two samples. We first used these methods with only the time series and then also provided the structured covariates as input by repeating them along the temporal dimension. Another baseline was Random Forests (RF) applied to top PCs extracted from time series and covariates. Finally, we ran RF on the structured covariates and then added features engineered by domain experts from time series data.

\subsection{Experimental Protocol}
\label{sec:protocol}

The data were split into training, validation, and test sets to perform hyper-parameter tuning and evaluation. We use a two-level sampling scheme. The outer loop consists of $M$ iterations, each randomly splitting the data 90\%/10\% for training+validation and test. Model selection is performed within the inner loop, with the 90\% set being randomly split once again 90/10 into training and validation sets. There are $N$ rounds of subsampling for training+validation, with the average performance over the validation set being used to select the best performing model. The hyper-parameters with the best validation accuracy are chosen, and the model with these parameters is trained on the 90\% training+validation dataset. The range of hyperparameters for each model is in Appendix~C, Table~\ref{tab:hyper}. The model is evaluated on the 10\% test set in the outer loop. We report the average accuracy over the $M$ test sets. In this work, we use $M=5$ and $N=10$. We tested the models described in the previous sections, with hyperparameters obtained by our 2-level model selection.

\section{Discussion of Results}
\label{sec:discussion}

\begin{table}[htbp]
  \centering
  \caption{Accuracy of the deep learning models on the benchmark datasets. The checkmarks indicate which types of data -- covariates or time series -- are used by each model.}
    \begin{tabular}{|l|l|l|l|l|}
    \hline
          &  Cov. & Time  & \textbf{CP Psoas} & \textbf{OAI} \\
          &  & series  & \textbf{Prediction} & \textbf{Progression} \\
    \hline
    Default predictor & - & - & 65\% & 63\% \\ \hline
    COV+RF & \checkmark &  - & 67\%  & 65\% \\ \hline
    Engineered Features + RF & \checkmark & \checkmark & 78\% $^{*}$ & 67\% \\ \hline
    PCA+COV+RF & \checkmark & \checkmark & 72\% & 67\% \\ \hline
    MKL  & - & \checkmark & 64\% & 67\% \\ \hline
    MKL+COV & \checkmark & \checkmark & 76\%  & 68\% \\ \hline
    DTW  & -  & \checkmark & 72\%  & 71\% \\ \hline
    DTW+COV & \checkmark & \checkmark & 72\%  & 71\% \\ \hline
    RNN/LSTM & - & \checkmark & 68\% & 69\% \\ \hline
    Multiresolution CNN & - & \checkmark & 75\% & 71\% \\ \hline
    Encoder + CNN & - & \checkmark & 75\%  & 68\% \\ \hline
    CNN  & - & \checkmark & 75\% & 68\% \\ \hline
    \textbf{LateFuse} & \checkmark & \checkmark & 77\% [BASE = CNN] & 70\% [BASE=LSTM] \\ \hline
    \textbf{ShortFuse} & \checkmark & \checkmark & \textbf{78}\% [BASE = CNN] & \textbf{74}\% [BASE=LSTM] \\
    \hline
    \end{tabular}%
  \label{tab:results} \\
$^{*}$ As obtained by \citep{schwartz2013predicting}.
\end{table}%

The two biomedical applications, osteoarthritis progression and cerebral palsy surgical outcome are described in detail in the next sections. The class imbalance is 63\% for osteoarthritis progression and 65\% for the Psoas prediction. The results are summarized in Table~\ref{tab:results}. For both applications, the RF models trained exclusively on covariates are the worst performers, indicating that time series should be used in the prediction. The results also show that ShortFuse, which introduces structured covariates in deep learning architectures built for the processing of time series data, is 3\% more accurate than past deep learning models and other methods for automatically learning time series representations. ShortFuse also matches or outperforms models trained by domain experts. We also find that ShortFuse outperforms LateFuse by 2-4\% -- the structured covariates have a greater impact in the representation learning if they are integrated into the network as opposed to being merged right before the prediction. ShortFuse also outperforms MKL, DTW and PCA+RF by 2-3\%, even when these methods use the structured covariates. Unlike in the case of deep learning models, providing covariates to MKL and DTW did not lead to significant increase in accuracy. The hypothesis space expressed by these models is not rich enough to explain the underlying connections between the time series and the structured data. 

\subsection{Forecasting osteoarthritis progression}
\label{sec:oai} 

Knee osteoarthritis (OA) is a leading cause of disability in older adults \citep{centers2009prevalence,guccione1994effects}, with 50\% of the population at risk of developing symptoms at some point in their life \citep{murphy2008lifetime}. Prevention, which could significantly reduce the burden of this incurable disease, hinges on a deeper understanding of modifiable risk factors, such as physical activity \citep{dunlop2014relation,lee2015sedentary}. Currently, clinicians lack the necessary evidence to make specific activity modification recommendations to patients.  Some studies have reported that physical activity is associated with an increased risk of knee OA \citep{lin2013physical,felson2013physical}. Others have reported either no association or opposite findings \citep{racunica2007effect,mansournia2012effect}. Current suggestions are not fine-tuned to patient demographics, medical histories, and lifestyles. Similar types of activities are expected to have different effects on patients with different joint alignment angles or different levels of systemic inflammation \citep{griffin2005role}. The interaction of these covariates with physical activity is thus important in predicting disease progression. In this example application, our task is to predict the progression of osteoarthritis, in terms of an objective measure of cartilage degeneration called Joint Space Narrowing (JSN).
 
We use a dataset of 1926 patients collected as part of the Osteoarthritis Initiative \citep{oainitiative17}, an ongoing 7-year longitudinal observational study on the natural progression of knee OA that monitored patients yearly, collecting medical histories, nutritional information, medication usage, accelerometer-based physical activity data, and other data from OA-related questionnaires. As part of the study, subjects had radiographs (X-rays) of their knees taken yearly and their activity monitored for one week. Activity time series were provided as activity counts (acceleration time steps per minute). X-ray data had been previously processed to extract the joint space width, or the distance between the thigh and shank bones, which is representative of cartilage thickness. As cartilage degenerates, the joint space becomes narrower. If the decrease in cartilage width is higher than 0.7mm per year, the disease is said to have progressed. We used covariates  from years 0-4 and physical activity time series from year 4 to predict whether the disease progressed from year 4 to year 6. The structured covariates include clinical features extracted from knee X-rays, while the time series data represents activity counts obtained over a week of monitoring. The human engineered features for this task are based on histograms for the activity data.

An RF model that featurizes the activity data using a histogram approach where features are activity totals for 10 bins of activity intensity levels obtains a 67\% classification accuracy, which is only slightly above random chance, once class imbalance is accounted for. The best base deep learning architecture is LSTM, which we found to perform well for the single-sequence, non-periodic in this application. ShortFuse with a hybrid LSTM obtains an accuracy of 74\%, a 7\% increase over the histogram and RF approach and a 5\% increase over a standard LSTM.

\subsection{Predicting the outcome of surgery in patients with cerebral palsy}
\label{sec:cp}

Cerebral palsy is a disorder of movement, muscle tone, and/or posture that is caused by damage that occurs to the immature, developing brain, most often before birth. The condition affects 500,000 people in the US (3.3 per 1,000 births), with 8,000 babies and infants diagnosed each year \citep{bhasin2006prevalence}. Automated tools are needed to aid treatment planning and predict surgical outcomes  given both the complexity of the disease (patients present with widely varying gait pathologies) and invasive nature of treatments, which include skeletal, muscular, and neural surgeries.

In this application, our task is to predict whether psoas lengthening surgery (a procedure to address a tight or overactive muscle in the pelvic region) will have a positive outcome. As in previous work \citep{truong2011evaluation}, we define a positive outcome as (1) an improvement of more than 5 points in Pelvis and Hip Deviation Index (PHiDI), which is a gait-based measure of dysfunction of motion of the pelvis and hip during walking, or (2) a post-surgical Gait Deviation Index of more than 90, which indicates that the subject's gait pattern is within one standard deviation of a typically-developing child. The time series in the data are joint angles obtained during the subject's gait cycle from motion capture using markers. The computation of the human engineered features requires domain expertise such as knowledge of the stances in the gait cycle (i.e., whether the foot of the limb of interest is in contact with the ground or not). 

The current state of the art uses an RF model trained on clinical information as well as engineered features, which has an accuracy of 78\% \citep{schwartz2013predicting}. The best performing deep learning architecture is the deep CNN, possibly because the gait time series consists of multiple (15) sequential variables representing joint angles, which all have a shape that does not vary considerably across subjects. ShortFuse improves over the best deep learning model by 3\%, matching the performance of the model trained on human engineered features and covariates, thus obviating the need for human designed features.

\section{Conclusions}

We introduce a new method, called ShortFuse, that incorporates structured covariates into time series deep learning, which can improve performance over current state-of-the-art models. The key contribution over previous work is that the covariates have a direct effect on the representations that are learned, which, as shown in our example, leads in more accurate models. Results indicate that the structured covariates have a greater impact on the representation learning if they are integrated into the network early as opposed to being merged right before the final layer. We have also outperformed other standard baselines, even when the baselines use covariates. For example, for the cerebral palsy task, MKL with covariates obtains 76\% accuracy, whereas ShortFuse obtains 78\%.

ShortFuse obtains 3\% improvement over all other approaches in forecasting osteoarthritis-related cartilage degeneration, 3 years in advance. This is crucial in supporting clinicians in making informed recommendations for patients who present with joint pain. 
For surgery outcome prediction in cerebral palsy patients, we outperformed or matched the state-of-the-art, at the same time eliminating the need for painstaking feature engineering.

% ACKNOWLEDGEMENTS
 \acks{We would like to acknowledge support from the National Institutes of Health (U54 EB020405). Many thanks to Dr. Michael Schwartz at the University of Minnesota and Gillette Children's Specialty Healthcare Hospital for providing the cerebral palsy data. The OAI is a public-private partnership comprised of five contracts (N01-AR-2-2258; N01-AR-2-2259; N01-AR-2-2260; N01-AR-2-2261; N01-AR-2-2262) funded by the National Institutes of Health, a branch of the Department of Health and Human Services, and conducted by the OAI Study Investigators. Private funding partners include Merck Research Laboratories; Novartis Pharmaceuticals Corporation, GlaxoSmithKline; and Pfizer, Inc. Private sector funding for the OAI is managed by the Foundation for the National Institutes of Health. This manuscript was prepared using an OAI public use data set and does not necessarily reflect the opinions or views of the OAI investigators, the NIH, or the private funding partners. We also thank Prof. David Sontag for his helpful suggestions.}

\newpage
\appendix
\paragraph{Appendix A. List of hybrid layers} \
\label{app:A}

\begin{table}[h]
  \centering 
  \caption{List of ShortFuse hybrid layers and connections with standard layers.} 
  \begin{tabular}{|l|l|}\hline
    \textbf{Standard Layer} & \textbf{Hybrid Layer} \\ \hline
    Convolution 1D & Covariates provided to convolutions along temporal dimension. \\ \hline 
    Concolution 2D & Interleave covariates to obtain a sequence of the same periodicity \\ 
                             & and size as the time series data. \\ \hline 
    Fully Connected & Covariates inputted to each one of the fully connected cells. \\ \hline 
    RNN/LSTM & Use the structured covariates as part of additive terms in the \\
                             & computation of the LSTM parameters. \\ \hline 
  \end{tabular}
  \label{tab:hybrids} 
\end{table}

\paragraph{Appendix B. Figures of deep learning models} \
\label{app:B}

\begin{figure}[h]
  \centering 
  \begin{subfigure}[b]{0.32\textwidth}
        \centering
        \includegraphics[width=\textwidth]{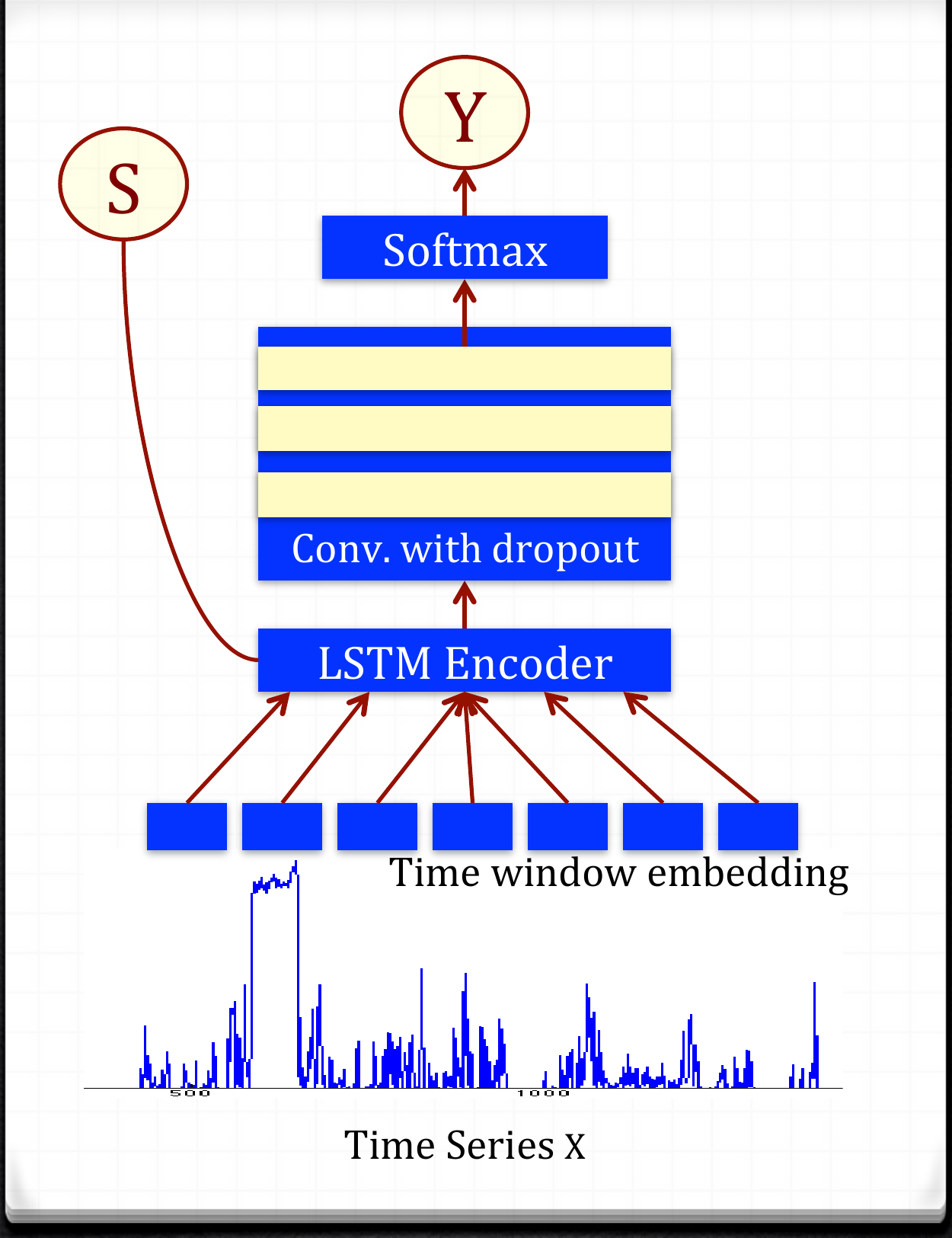}
        \caption{LSTM with embeddings.}
         \label{fig:4a}
    \end{subfigure}%
\qquad
  \begin{subfigure}[b]{0.48\textwidth}
        \centering
        \includegraphics[width=\textwidth]{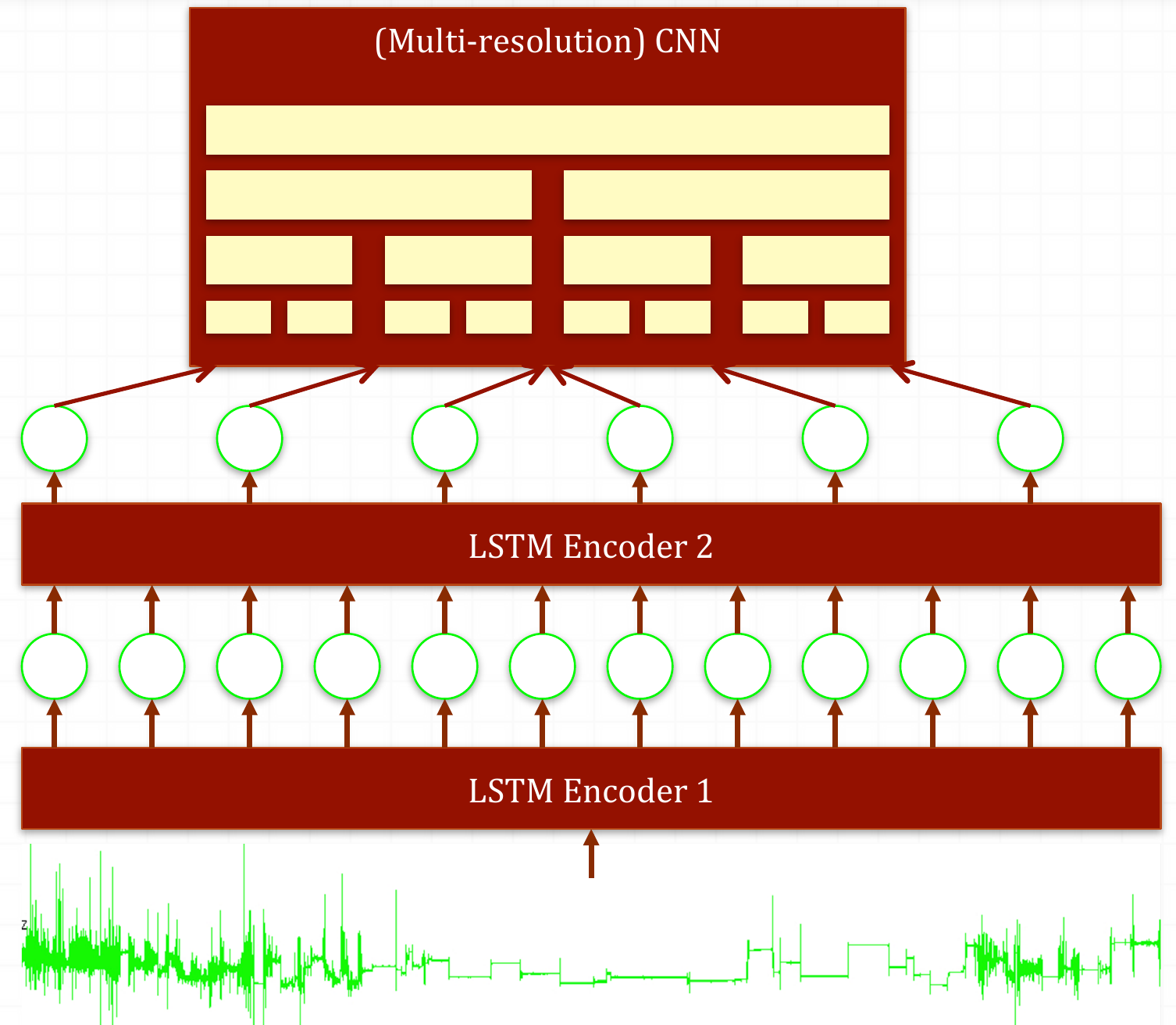} 
        \caption{Encoder + CNN.}
        \label{fig:4b}
    \end{subfigure}%
    \caption{Deep learning candidate models.}
\label{fig:LSTMmodel}
\end{figure}

\paragraph{Appendix C. Model parameters} \
\label{app:C}

\begin{table}[h]
  \centering
  \caption{Hyperparameters used in the model training and the models they apply to.}
    \begin{tabular}{|r|r|r|}
    \hline
    \textbf{Hyperparameter} & \textbf{Model to which it applies} & \textbf{Parameter range for search} \\ \hline
    Learning rate & RNN / LSTM / all CNN models & 0.001 - 0.003 \\ \hline
    Dropout & RNN / LSTM / all CNN models & 0.0 - 0.5 \\ \hline
    Embedding size & LSTM  & 16 - 64 \\ \hline
    Number of filters & all CNN models & 3-13 \\ \hline
    Number of layers & all CNN models & 1-10 \\ \hline
    Resolutions & Multiresolution CNN & 256 - 128 - 64 - 32 -16 \\ \hline
    Kernel & Multiple Kernel Learning & RBF \\ \hline
    Number of trees & Random Forests & 10 - 1000 \\ \hline
    \end{tabular}%
  \label{tab:hyper}%
\end{table}%
\end{document}